\documentclass[11pt]{article}
\usepackage{eacl2017}
\usepackage[utf8]{inputenc}
\usepackage{times}
\usepackage{url}
\usepackage{latexsym}
\usepackage{amssymb}
\usepackage{amsmath}
\usepackage{subfig}
\usepackage{graphicx}
\usepackage{booktabs}
\usepackage{url}
\usepackage{color}
\usepackage{hyperref} 

\eaclfinalcopy 

\title{Sentence Segmentation in Narrative Transcripts from Neuropsychological Tests using Recurrent Convolutional Neural Networks}

\author{{\begin{tabular}{ccc}
Marcos Vinícius Treviso 	& 	Christopher Shulby 			& 	Sandra Maria Aluísio 		\\
{\tt marcostreviso@usp.br}	&	{\tt cshulby@icmc.usp.br} 	& 	{\tt sandra@icmc.usp.br} 	\\
\multicolumn{3}{c}{\textnormal{Interinstitutional Center for Computational Linguistics (NILC)}}\\
\multicolumn{3}{c}{\textnormal{Institute of Mathematical and Computer Sciences}} \\
\multicolumn{3}{c}{\textnormal{University of São Paulo}} \\
\end{tabular}}}

\date{}

\begin{document}
\maketitle
\begin{abstract}
Automated discourse analysis tools based on Natural Language Processing (NLP) aiming at the diagnosis of language-impairing dementias generally extract several textual metrics of narrative transcripts. However, the absence of sentence boundary segmentation in the transcripts prevents the direct application of NLP methods which rely on these marks to function properly, such as taggers and parsers. We present the first steps taken towards automatic neuropsychological evaluation based on narrative discourse analysis, presenting a new automatic sentence segmentation method for impaired speech. Our model uses recurrent convolutional neural networks with prosodic, Part of Speech (PoS) features, and word embeddings. It was evaluated intrinsically on impaired, spontaneous speech, as well as, normal, prepared speech, and presents better results for healthy elderly (CTL) ($F_1$ = 0.74) and Mild Cognitive Impairment (MCI) patients ($F_1$ = 0.70) than the Conditional Random Fields method ($F_1$ = 0.55 and 0.53, respectively) used in the same context of our study. The results suggest that our model is robust for impaired speech and can be used in automated discourse analysis tools to differentiate narratives produced by MCI and CTL.
\end{abstract}

\section{Introduction}

Mild Cognitive Impairment (MCI) has recently received much attention, as it may represent a pre-clinical state of Alzheimer's disease (AD).
MCI can affect one or multiple cognitive domains (e.g. memory, language, visuospatial skills and the executive function); the kind that affects memory, called amnestic MCI, is the most frequent and that which most often converts to AD \cite{Janoutova:2015:NIPH}. As dementias are chronic progressive diseases, it is important to identify them in the early stages, because early detection yields a greater chance of success for non-pharmacological treatment strategies such as cognitive training, physical activity and socialization \cite{Art:Teixeira:2012:Non}.
The definition of MCI diagnostic criteria is conducted mainly by the cognitive symptoms presented by patients in standardized tests and by functional impairments in daily life \cite{Art:Mckhann:2011:DiagnosisDA}. Difficulties related with narrative discourse deficits (e.g. repetitions or gaps during the narrative) may lead an elderly individual to look for a specialist. Narrative discourse is the reproduction of an experienced episode (necessarily evoking memory), respecting temporal and causal relations among events. Although MCI is clinically characterized by episodic memory deficits, language impairment may also occur. 

Certain widely used neuropsychological tests require patients to retell or understand a story. This is the case of the logical memory test, where one reproduces a story after listening to it. The higher the number of recalled elements from the narrative, the higher the memory score \cite{Book:Wechsler:1997:WLM,Book:Bayles:1991:ABCD,Art:Morris:2006:LWW}.
However, the main difficulties in applying these tests are: (i) time required, since it is a manual task; and (ii) the subjectivity of the clinician.
Therefore, automatic analysis of discourse production is seen as a promising solution for MCI diagnosis, because its early detection ensures a greater chance of success in addressing potentially reversible factors \cite{Art:Muangpaisan:Prevalence:2012}. 
Since discourse is a natural form of communication, it favors the observation of the patient’s functionality in everyday life. Moreover, it provides data for observing the language-cognitive skills interface, such as executive functions (planning, organizing, updating and monitoring data).  

With regard to the Wechsler Logical Memory (WLM) test,  the original narrative used is short, allowing for the use of Automatic Speech Recognition (ASR) output even without capitalization and sentence segmentation, as shown by Lehr et al.~\shortcite{Inp:Lehr:2012:IS} for English. They based their method on automatic alignment of the original and patient transcripts in order to calculate the number of recalled elements.

The evaluation of narrative discourse production from the standpoint of linguistic impairment is an attractive alternative as it allows for linguistic microstructure analysis, including phonetic-phonological, morphosyntactic and semantic-lexical components, as well as semantic-pragmatic macrostructures. Automated discourse analysis tools based on Natural Language Processing (NLP) resources and tools aiming at the diagnosis of language-impairing dementias via machine learning methods are already available for the English language \cite{Art:Fraser:2015:linguistic,Inp:Yancheva:2015:Predict,Art:Roark:2011:Spoken} and also for Brazilian Portuguese (BP)  \cite{Art:Aluisio:2016:eval}. The latter study used a publicly available tool, Coh-Metrix-Dementia\footnote{\url{http://143.107.183.175:22380/}}, to extract 73 textual metrics of narrative transcripts, comprising several levels of linguistic analysis from word counts to semantics and discourse.
However, the absence of sentence boundary segmentation in transcripts prevents the direct application of NLP methods that rely on these marks in order for the tools to function properly. To our knowledge, only one study evaluating automatic sentence segmentation in English transcripts of elderly aphasic exists \cite{Inp:Fraser:2015:segmentation}.

The purpose of this paper is to present our method, DeepBond, for automatic sentence segmentation of spontaneous speech of healthy elderly (CTL) and MCI patients. Although it was evaluated for BP data, it can be adapted to other languages as well.

\section{Related Work}

The sentence boundary detection task has been treated by many researchers. Liu et al. \shortcite{Art:Liu:2006:sbd} investigated the imbalanced data problem, since there are more non-boundary words than not; their study was carried out using two speech corpora: conversational telephone and broadcast news, both for English. 

More recent studies have focused on Conditional Random Field (CRF) and Neural Network models. Wang et al. \shortcite{Art:Wang:2012:dynamic} and  Hasan et al. \shortcite{Inp:Hasan:2014:multi} use CRF based methods to identify word boundaries in speech corpora datasets, more specifically on English broadcast news data and English conversational speech (lecture recordings), respectively. Khomitsevich et al. \shortcite{Inp:Khomitsevich:2015:combining}, similar to our work, used a combination of two models, one based on Support Vector Machines to deal with prosodic information, and other based on CRF to deal with lexical information. They combine the two models using a logistic regression classifier. 

Xu et al. \shortcite{Inp:xu:2014:deep} uses a combination of CRF and a Deep neural network (DNN) to detect sentence boundaries on broadcast news data. Che et al. \shortcite{Art:Che:2016:convolutional} uses two different convolutional neural network (CNN), one which moves in only one dimension and another which moves in two. They achieved good results on a TED talks dataset. Tilk and Alum{\"a}e \shortcite{Inp:Tilk:2015:lstm} use a recurrent neural network (RNN) with long short-term memory units to restore punctuation in speech transcripts from broadcast news and conversations.

Although there are proposed methods for sentence segmentation of Portuguese datasets \cite{Inp:Silla:2004:analysis,Phd:Batista:2011:recovering,Inc:Lopez:2015:Experiments}, none of them are used for transcriptions produced in a clinical setting for the elderly with dementia and related syndromes. The study most similar to our scenario is \cite{Inp:Fraser:2015:segmentation}, which proposes a segmentation method for aphasic speech based on lexical, PoS and prosodic features using tools and a generic acoustic model trained for English. Their approach is based on a CRF model, and the best results for this study were obtained for non-spontaneous broadcast news data.

Our method uses recurrent convolutional neural networks with prosodic, PoS features, and also word embeddings and was evaluated intrinsically on impaired, spontaneous speech and normal, prepared speech. Although DNNs have already been used for this task, our work was the first, to the best of our knowledge, to evaluate them on impaired speech.



\section{Datasets}

A total of 60 participants from a research project on diagnostic tools for language impaired dementias produced narratives used to evaluate our method. Two datasets were used to train our model (Sections 3.1 and 3.2). As a preprocessing step we have removed capitalization information and in order to simulate high-quality ASR, we left all speech disfluences intact. Demographic information for participants in our study is presented in Table~\ref{tab:demographic-table}. A third dataset was used in robustness tests (Section 3.3).


\begin{table}[!htb]
\begin{center}
\begin{tabular}{lccc}
\toprule 
\bf Info & \bf CTL & \bf MCI & \bf AD 		\\ 
\midrule
Avg. Age 			& 74.8 & 73.3 & 78.2 	\\
Avg. Education		& 11.4 & 10.8 & 8.6 	\\
No. of Male/Female 	& 4/16 & 6/14 & 10/10 	\\
\bottomrule
\end{tabular}
\end{center}
\caption{\label{tab:demographic-table} Demographic information of participants in the Cinderella dataset. The Avg. Education is given in years.}
\end{table}	

\subsection{The Cinderella Dataset}
The Cinderella dataset consists of spontaneous speech narratives produced during a test to elicit narrative discourse with visual stimuli, using a book consisting of sequenced pictures based on the Cinderella story. In the test, an individual verbally tells the story to the examiner based on the pictures. The narrative is manually transcribed by a trained annotator who scores the narrative by counting the number of recalled propositions.

This dataset consists of 60 narrative texts from BP speakers, 20 controls, 20 with AD, and 20 with MCI, diagnosed at the Medical School of University of São Paulo and also used in
Aluísio et al.~\shortcite{Art:Aluisio:2016:eval}. 
Counting all patient groups, this dataset has an audio duration of 4h and 11m, an average of $1843/60 = 30.72$ sentences per narrative, and sentence averages of $23807/1843 = 12.92$ words. AD narratives were only used for training the lexical model.

\subsection{The Brazilian Constitution Dataset}
This dataset was made available by the LaPS (Signal Processing Laboratory) at the Federal University of Pará \cite{Art:Batista:2013:avancos}, and is composed of articles from Brazil's 1988 constitution, in which the speech is prepared and read. Each file has an averages 30 seconds. 

A preprocessing step removed lexical tips which indicate the beginning of the articles, sections and paragraphs. This removal was carried out on both the transcripts and audio. In addition, we separated the new dataset organized by articles, totaling 357 texts. Then, we marked the end of each article and paragraph and inserted punctuation at the end. Titles and chapters have been ignored during this process. We randomly selected 60 texts from this dataset, forcing only the condition that the number of sentences of each text sentence was greater than 12. We refer to the large dataset as Constitution L, and the dataset with the 60 texts as Constitution S.

The average number of sentences in each text of Constitution L is $2698/357 = 7.56$, and the average size of these sentences have $63275/2698 = 23.45$ words while Constitution S has on average $1409/60 = 23.48$ sentences, and these sentences average $30521/1409 = 21.66$ words. The total audio duration of Constitution L is 7h 39m, and Constitution S is 3h 43m.

\begin{figure*}[!htb]
  \center
  \includegraphics[width=0.98\textwidth]{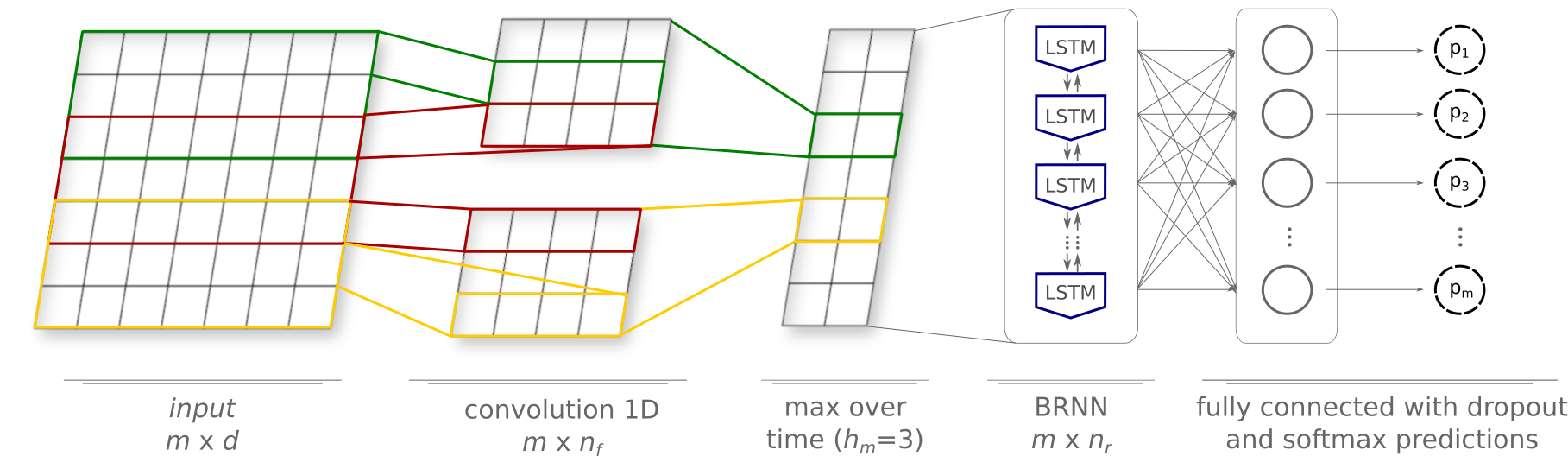}
  \caption{Architecture of the RCNN for both lexical and prosodic model.}
  \label{fig:architecture}
\end{figure*}

\subsection{The Dog Story Dataset}
The Dog Story dataset is available from the BALE (Battery of Language Assessment in Aging, in English) instrument, described in \cite{Phd:Jeronimo:2016:produccao}. It is composed of transcriptions from the narrative production test based on the presentation of a set of seven pictures telling a story of a boy who hides a dog that he found on the street \cite{Book:LeBoeuf:1976:raconte}.
This battery was chosen because its aim is to allow for its administration to elderly people who are illiterate and/or of low educational level, who represent the majority of the aged sample assisted by the public health system in Brazil.

This dataset consists of $10$ narratives transcripts ($6$ CTL and $4$ MCI), where the average number of sentences and the average size of the sentences are 16.60 and 6.58, respectively. When compared with the Cinderella dataset, the dataset is composed of less sentences and the sentences have fewer words on average.

\section{Features}

\subsection{Lexical features}

We divide our lexical features into two groups: PoS features and word embeddings, where every word is represented in a high dimensionality continuous vector. 

The PoS features where extracted using a BP morphosyntatic tagger called nlpnet\footnote{\url{nilc.icmc.usp.br/nlpnet/}} trained on a revised version of the Mac-Morpho corpus \cite{Art:Fonseca:2015:evaluating}, which contains a set of 25 tags.


The word embeddings used in this work have 50 dimensions and were trained by Fonseca et al. \shortcite{Art:Fonseca:2015:evaluating} 
with articles from the BP version of Wikipedia and a large journalistic corpus with articles from the news site G1\footnote{\url{g1.globo.com/}}, totaling 240 million tokens and a vocabulary of 160,270 words. All of these tokens were made lowercase and trained with a neural language model described in \cite{Art:Collobert:2011:natural}. 

\subsection{Prosodic features}

We used three prosodic features: F0, intensity and duration which were extracted at the phonetic level using PRAAT \cite{boersma2002praat} from forced alignment output. Alignment was done using using the HTK toolkit \cite{young2002htk} with clean speech corpora and a pronunciation dictionary phonetically transcribed by Petrus \cite{serrani2015ambiente} and augmented by our rule-based algorithm to insert multiple pronunciations, rendering a suitable model for ASR. The features were calculated for the first, last, penultimate and antepenultimate vowels of each word and pauses. These vowels were chosen based on knowledge of the BP which typically exhibits stress on the penultimate vowel, with notable patterns observed for final vowel stressing, for example words ending in ``i'' (``Barueri'') or a nasal consonant (``Renan''), and the antepenultimate vowel (usually indicated by a stress diacritic) like ``helicóptero'' (``helicopter''), ``espírito'' (``spirit'') and ``árvore'' (``tree'').  Also, Portuguese, like most western languages, distinguishes sentence types by rising and falling pitch patterns, giving the listener a clue as to whether the speaker has finished a sentence or not. Pause duration was also calculated since the length of a pause can be indicative of the presence of a punctuation mark \cite{beckman1997guidelines}.

\section{Model description}

To automatically extract features from the input and also deal with the problem of long dependencies between words, we propose a model based on recurrent convolutional neural networks (RCNN), which was inspired by the work of Lai et al.~\shortcite{Art:Lai:2015:rcnn}. The architecture of our model can be seen in Figure~\ref{fig:architecture}. First, we show how to prepare the input for the network, then we go through the networks layers and describe the training procedure, finally, we discuss the experimental settings.


\subsection{Input preparation}

In our approach, the input to the network is a transcribed narrative which is categorized as CTL (healthy elderly individuals) and MCI (MCI patients). The narratives contain a sequence of words $w_1, w_2, \dotsc, w_m$. Each word is annotated with a label, to indicate whether it precedes a boundary $(y = B)$ or not $(y = NB)$. We do not make a distinction between punctuation marks, so a boundary is defined as a period, exclamation mark, question mark, colon or semicolon. With this approach, we can see this task as a  binary classification problem. 

\subsection{Representation}

Our input contains transcribed narratives with $m$ words in it. We represent the narrative $i$ as $X_i \in \mathbb{R}^{m \times n}, X_i = \{x_1, x_2, \dotsc, x_{m \times n}\}$, where $n$ is the number of features. We represent the boundaries as $Y_i \in \mathbb{R}^2, Y_i = \{0, 1\}$, where $0$ stands for $NB$ and $1$ denotes $B$. 
Our final model consists of a combination of two models. The first model is responsible for treating only lexical information, while the second treats only prosodic information. Both models have the same architecture shown in Figure~\ref{fig:architecture}. This strategy is based on the idea that we can train the lexical model with even more data, since textual information is easily found on the web. In order to obtain the most probable class $y$ for the $w_j$ word, a linear combination was created between these two models, where one receives the weighted complement of the other:
\begin{equation} \label{eq:linearcomb}
\alpha \cdot P_{lexical} (y \, | \, w_j) + (1 - \alpha) \cdot P_{prosodic} (y \, | \, w_j)
\end{equation}
Then, the most probable class is the one that maximizes the linear combination from previous equation.
\subsubsection{Embedding layer}
The data input for the lexical model is divided into two features: word embeddings with dimensions $|e_w|$, and the PoS tags with dimensions $|e_t|$. Given a word $w$, the respective embedding $e_w \in E_{word}$ is fetched and concatenated with the word’s PoS vector $e_t \in E_{tag}$, thus obtaining a new vector size $d = |e_w| + |e_t|$. Out of vocabulary words share a single and randomly generated vector that represents an unknown word.

In the prosodic model we directly feed information about pitch, intensity and duration from the first, last, penultimate and ante-penultimate vowels of each word. Moreover, we feed the information about pause duration after each word, where  duration of zero seconds denotes no pause. Therefore, for the prosodic model, we have a vector with dimensions $d = 4 \cdot 3 + 1 = 13$.

\subsubsection{Convolutional and pooling layer}

Once we have a matrix formed by the features of the words in the text, the convolutional layer receives it, which, in turn, is responsible for the automatic extraction of $n_f$ new features depending on $h_c$ neighboring words \cite{Inp:Kim:2014:convolutional}. The convolutional layer produces a new feature $c_j$ by applying a filter $W \in \mathbb{R}^{h_c \cdot d}$ to a window of $h_c$ words $x_{j-h_c+1:j}$ in a sentence with length $m$:
\begin{equation} \label{feature_map} 
c_j = f(W x_{(j-h_c+1) : j} + b) , \quad h_c \leq j \leq m
\end{equation}

Where $b \in \mathbb{R}$ represents a bias term and $f$ is a non-linear function. 

Our convolutional layer simply moves one dimension vertically, making one step at a time, which gives us $m - h_c + 1$ generated features. Since we want to classify exactly $m$ elements, we add $p = \lfloor h_c / 2 \rfloor$ zero-padding on both sides of the text. Applying this strategy for each entry $x_j$ yields the complete feature map $c \in \mathbb{R}^{(m - h_c + 1) + 2 \cdot p}$.

In addition, we apply a max-pooling operation over time, looking at a region of $h_m$ elements to find the most significant features:
\begin{equation}
\hat{c} = \max\limits_{1 \leq j \leq m} \{ c_{(j-h_m+1):j} \}
\end{equation}

\subsubsection{Recurrent layer}

\begin{figure}[!htb]
	\centering
    \includegraphics[width=0.425\textwidth]{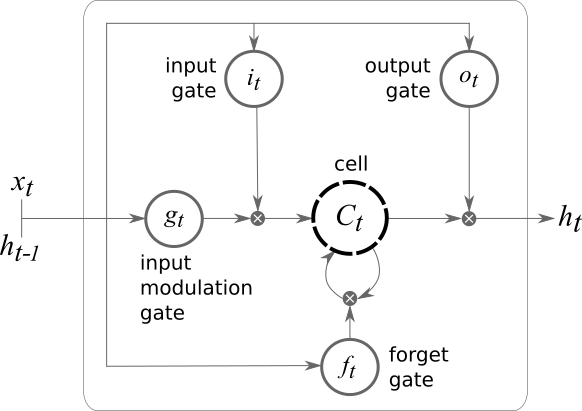}
  \caption{Diagram of a LSTM memory cell.}
  \label{fig:lstm}
\end{figure}

The new features extracted are fed into a recurrent bidirectional layer which has $n_r$ units. A recurrent layer is able to store historic information by connecting the previous hidden state with the current hidden state at a time $t$. The values in the hidden and output layers are computed as follows:
\begin{align}
h_t & = f (W_x x_t + W_h h_{t-1} + b_h) \label{eq:rnn_ht} \\
y_t & = g (W_y h_t + b_y) \label{eq:rnn_yt}
\end{align}

where $W_x$, $W_h$, and $W_y$ are the connection weights, $b_y$ and $b_h$ are bias vectors, and $f$ and $g$ are non-linear functions. Here, we use a special unit known as Long Short-Term Memory (LSTM) \cite{Art:Hochreiter:1997:lstm}, which is able to learn over long dependencies between words by a purpose-built memory cell. Figure~\ref{fig:lstm} shows a single LSTM memory cell. 

The LSTM updates for time steps $t$ are done as described by Jozefowicz et al. \shortcite{Art:Jozefowicz:2015:empirical}, which is a slight simplification of the one described by Graves and Jailty \shortcite{Inp:Graves:2014:towards}, where the memory cell is implemented as follows:

\begin{align*}
i_t &= \sigma ( W_{xi} x_t + W_{hi} h_{t-1} + b_i ) 
\\
f_t &= \sigma ( W_{xf} x_t + W_{hf} h_{t-1} + b_f ) 
\\
o_t &= \sigma ( W_{xo} x_t + W_{ho} h_{t-1} + b_o ) 
\\
g_t &=   tanh ( W_{xc} x_t + W_{hc} h_{t-1} + b_c)
\\
c_t &=  f_t \odot c_{t-1} + i_t \odot g_t 
\\
h_t &= o_t \odot tanh(c_t)
\end{align*}

where $\sigma(z) = 1 / (1 + e^{-z})$ is the sigmoid function, $h_t \in \mathbb{R}^{n_r}$ is the hidden unit, $i_t \in \mathbb{R}^{n_r}$ is the input gate, $f_t \in \mathbb{R}^{n_r}$ is the forget gate, $o_t \in \mathbb{R}^{n_r}$ is the output gate, $g_t \in \mathbb{R}^{n_r}$ is the input modulation gate, and $c_t \in \mathbb{R}^{n_r}$ is the memory cell unit, which is the summation of the previous memory cell modulated by the forget gate $f_t$, and a function of the current input with previous hidden state modulated by the input gate $i_t$.

As in Graves and Jaitly \shortcite{Inp:Graves:2014:towards}, we used the features by looking at forward states and backward states. This kind of mechanism is known as a bidirectional neural network (BRNN), since it learns weights based on both past and future elements given a timestep $t$. In order to implement the BRNN, we reversed the sentences as a trick before we fed them to a regular LSTM layer, doubling the number of weights used in the recurrent layer. The output from this layer is the summation of the forward output with backward output:
\begin{equation}
y_t = \overleftarrow{y_t} + \overrightarrow{y_t}
\end{equation}

With a bidirectional LSTM layer, we are able to explore the principle that words nearby have a greater influence in classification, while considering that words farther away can also have some impact. This often happens, for example, in the case of question words and conjunctions: por que (“why”); qual (“which”); quem (“who”); quando (“when”), etc.

\subsubsection{Fully connected layer}

After the BRNN layer, dropout is used to prevent co-adaptation of hidden units during forward-backpropagation, where we ignore some neurons meaning to reduce the chance of overfitting the model \cite{Art:Srivastava:2014:dropout}.

The last layer receives the output from the BRNN in each timestep and passes them trough a fully connected layer, where the softmax operation is calculated, giving us the probability of whether or not the word precedes a boundary: 
\begin{equation}
\hat{y}_t = softmax(W y_t + b)
\end{equation}

Where $W \in \mathbb{R}^{n_r \times 2}$ is a matrix of weights, $b \in \mathbb{R}^{n_r} $ is a bias vector, and softmax is defined as:
\begin{equation}
s_j(z) = \frac{e^{z_j}}{\sum_{k=1}^{K}e^{z_k}} , \quad \textnormal{for} \, j = 1, 2, \dots, K
\end{equation}

\subsection{Training}

We define all of the parameters to be trained as $\theta$.
\begin{align}
\theta = \big\{ & E_{word},\, E_{tag},\, 
				W^{(c)},\, b^{(c)},\,W^{(f)}, \nonumber \\
				& b^{(f)},\, \overleftarrow{W}^{(r)},\, \overleftarrow{b}^{(r)},\,
                \overrightarrow{W}^{(r)},\, \overrightarrow{b}^{(r)}
 \big\}
\end{align}

Where $E_{word} \in \mathbb{R}^{|V| \times |e_w|}$ is the lookup table for the word embeddings, $E_{tag} \in \mathbb{R}^{|V_{tag}| \times |e_t|}$ is the lookup table for PoS tags, and $|V|, |V_{tag}|$ represents the size of the vocabulary for word embeddings and PoS tags, respectively. 

For the convolutional layer: the weights $W^{(c)} \in \mathbb{R}^{n_f \times h_c \cdot d}$ and the bias vector $b^{(c)} \in \mathbb{R}^{n_f}$. 

For the fully connected layer: the weights matrix $W^{(f)} \in \mathbb{R}^{n_r \times 2}$ and the bias vector $b^{(f)} \in \mathbb{R}^{n_r}$. 

For the BRNN layer we divide the set of parameters from BRNN into two sets. Those from the forward pass and backward pass. Each set contains the weights for an input $W^{(r)}_x \in \mathbb{R}^{n_r \times n_f}$, the weights for previous hidden states $W^{(r)}_h \in \mathbb{R}^{n_r \times n_r}$, and the bias vectors $b^{(r)} \in \mathbb{R}^{n_r}$ for all gates ($i, f, o, g$). Additionally, we have the weights for an output in a timestep $W^{(r)}_y \in \mathbb{R}^{n_r \times n_r}$ and a bias vector $b_y \in \mathbb{R}^{n_r}$.

We define the loss function $\mathcal{L}$ as categorical cross-entropy \cite{Book:Murphy:2012:ml}, shown in the equation below, which aims to minimize the negative log likelihood in relation to the weights. Since we have an unbalanced class problem, we give different weights for each class, where the weight of the minority class ($B$) is greater than that of the majority ($NB$).
\begin{equation}
\mathcal{L}(y, \hat{y}) = - \sum\limits_{i} y_i \, log(\hat{y}_i) \, cw_{y_i}
\end{equation}

Where $y$ are our real targets, $\hat{y}$ are our predictions, and $cw$ are the class weights for $\ell = B$ and $\ell = NB$, calculated as follows:
\begin{equation}
cw_{\ell} = \frac{|y|}{2 \cdot |y = \ell|}
\end{equation}

We minimize the loss function with respect to all weights $\theta \mapsto \mathcal{L}$ by using RMSProp algorithm \cite{Art:Tieleman:2012:lecture} with backpropagation to compute the gradients $\nabla \mathcal{L}$. The update step for a timestep $t$ is made by normalizing the gradients by an exponent moving at an average $r_t$:
\begin{align}
r_t &= \gamma r_{t-1} + (1 - \gamma) \nabla \mathcal{L}(\theta_t) ^ 2 \\ \nonumber \\
\theta_{t+1} &= \theta_t - \eta \frac{\nabla \mathcal{L}(\theta_t)}{\sqrt{r_t} + \epsilon}
\end{align}

Where $\eta$ is the learning rate and $0 < \gamma < 1$ is the forgetting factor.

\subsection{Experiment settings}

We break the text in tokens delimited by spaces. We do not remove stopwords from the texts, since they can be important features for our domain. 


We ran a 5-fold cross-validation for the group being analyzed (CLT or MCI),  which leaves about 10\% of the data for testing, the rest for training.

The weight matrix for tag embeddings $E_{tag}$ was generated randomly from a gaussian distribution scaled by fan in + fan out \cite{Inp:Glorot:2010:understanding}. Both embeddings matrix $E_{word}$ and $E_{tag}$ were adjusted during training. We follow previous studies on sentence boundary detection to set the network hyper-parameters \cite{Inp:Tilk:2015:lstm,Art:Che:2016:convolutional}. The values for each parameter are shown in Table~\ref{tab:hyperparams}.
\begin{table}[!htb]
\begin{center}
\begin{tabular}{llcc}
\toprule
\bf Var. 			& \bf Parameter		& \bf Lexical & \bf Prosodic	\\ 
\midrule
$|e_w|$ 		 	 & Word emb. size  					& 50				& - 	\\
$|e_t|$				 & Tag emb. size 					& 10  				& - 	\\
$n_f$				 & Conv. filters					& 100				& 8 	\\
$h_c$				 & Filter length					& 7					& 5		\\
$h_m$				 & Max-pool size					& 3					& 3		\\
$n_r$				 & Recurrent units					& 100				& 100 	\\
$\gamma$			 & Forget factor					& 0.9				& 0.9	\\	
$\eta$				 & Learning rate					& 0.001				& 0.001	\\
\bottomrule
\end{tabular}
\end{center}
\caption{\label{tab:hyperparams} RCNN Hyper-parameters.}
\end{table}	

We tried three different learning rate values $\eta \in \{0.01, 0.003, 0.001\}$ for both lexical and prosodic models, and found that 0.001 yielded best results. We trained our network over 20 epochs using a bucket strategy, which groups training examples in buckets of similar sentence size. 
Our implementation is based on Theano \cite{Inp:Bergstra:2010:theano}, a library that defines, optimizes and evaluates mathematical expressions in an effective way.

\section{Evaluation}

We evaluated our method intrinsically and also compared it with the method developed by Fraser et al. \shortcite{Inp:Fraser:2015:segmentation} for all of the datasets. We also performed robustness tests to indicate how well our method responds to both (i) test data that varies from Cinderella training data and (ii) train data that varies from Cinderella testing data.


If we classified all words as $NB$, our method would have an accuracy superior to 90\%. For this reason, we use the $F_1$ metric, which is defined as the harmonic mean between precision and recall. And since we are more interested in knowing whether our method correctly identifies the boundaries, we ignore the $NB$s and calculate $F_1$ only for the positive class ($B$).

\subsection{Results}

In this subsection, we evaluate the performance of our classifier (RCNN) for the Cinderella and Constitution datasets. Table~\ref{tab:results} summarizes the results.

\begin{table*}[!htb]
\small
\begin{center}
\begin{tabular}{lccccccccccccccc}
\toprule
\bf Features & \multicolumn{7}{c}{\bf Cinderella} & & \multicolumn{7}{c}{\bf Constitution} \\ 					
			   \cmidrule{2-8} 
               \cmidrule{10-16}
             
               & \multicolumn{3}{c}{\bf CTL} & & \multicolumn{3}{c}{\bf MCI} & & 
                 \multicolumn{3}{c}{\bf L} & & \multicolumn{3}{c}{\bf S} \\
               
               \cmidrule{2-4} 
               \cmidrule{6-8}
               \cmidrule{10-12}
               \cmidrule{14-16}
             
             & \it P & \it R & \it $F_1$ & & \it P & \it R & \it $F_1$ & &
               \it P & \it R & \it $F_1$ & & \it P & \it R & \it $F_1$ \\  
\midrule 
Baseline 		& 0.07 & 1.00 &  0.13  		& &  0.08 & 1.00 & 0.14   	 & &  0.03 & 1.00 & 0.07  	  & &  0.04 & 1.00 & 0.08  		\\
PoS 			& 0.36 & 0.82 &  0.50  		& &  0.32 & 0.83 & 0.46   	 & &  0.30 & 0.89 & 0.44  	  & &  0.29 & 0.79 & 0.42   	\\
Prosody 		& 0.20 & 0.59 &  0.30  		& &  0.19 & 0.58 & 0.29   	 & &  0.54 & 0.84 & 0.66  	  & &  0.48 & 0.85 & 0.61  		\\
Embeddings 		& 0.70 & 0.70 &  0.70  		& &  0.63 & 0.77 & 0.69   	 & &  0.60 & 0.63 & 0.63 	  & &  0.60 & 0.64 & 0.62   	\\
PoS + Pros. 	& 0.40 & 0.74 &  0.52  		& &  0.36 & 0.80 & 0.49   	 & &  0.52 & 0.91 & 0.66  	  & &  0.57 & 0.85 & 0.68   	\\
Emb. + PoS 		& 0.71 & 0.72 &  0.71  		& &  0.64 & 0.75 & 0.69   	 & &  0.64 & 0.72 & 0.68  	  & &  0.63 & 0.67 & 0.65   	\\
Emb. + Pros. 	& 0.71 & 0.74 &  0.72  	  	& &  0.64 & 0.77 & 0.70   	 & &  0.71 & 0.83 & 0.76  	  & &  0.74 & 0.81 & 0.77  		\\
All	 			& 0.72 & 0.76 &  \bf 0.74 	& &  0.66 & 0.74 & \bf 0.70  & &  0.77 & 0.82 & \bf 0.79  & &  0.76 & 0.85 & \bf 0.80  	\\   
\bottomrule
\end{tabular}
\end{center}
\caption{\label{tab:results} $F_1$ for boundary class for each feature set on Cinderella and Constitution data using our method.}
\end{table*}


From Table~\ref{tab:results} we can see that our approach presents better results for the Constitution dataset than Cinderella. This may be related to the text quality, as the Cinderella transcripts presents many disfluences, characteristic of spontaneous speech. As expected, results for CTL were higher than for MCI, since CTL narratives contain less disfluencies. 
Another important observation is that our method performs much better than the baseline. Where the baseline represents the results for a classifier that predicts all words as $B$.  
The Constitution results show us that traditional machine learning techniques used in NLP can be applied to this scenario, since the differences in the Cinderella data are few. Another reason that supports this statement is that $F_1$ results from related studies on sentence boundary detection based on well-written texts are between $0.7$ and $0.8$ for two classes  \cite{Art:Wang:2012:dynamic,Inp:Khomitsevich:2015:combining,Inp:Tilk:2015:lstm,Art:Che:2016:convolutional}.
When we compare the Constitution size relation we find out that corpus size is not greatly affected by the results, since the results for Constitution S were slightly better than for Constitution L. We think that, even with less data, our method performs better on Constitution S because of the distribution of sentence quantity in the dataset, where Constitution S has an average of 23.48 sentences per text, while Constitution L has an average of only 7.56 sentences per text.    

We also evaluated the performance of different feature sets with our datasets.  Embeddings have a great impact on both datasets. The PoS information was influential on both datasets, but by a small margin, since it has a small difference when used with embeddings (0.01) on the Cinderella, and (0.03) Constitution data. This tells us that embeddings already bring enough morphosyntactic information. It is evident that the weight of the prosodic features is higher on Constitution, which is based on prepared speech, than in Cinderela. This result is consistent with those found by Kolár et al. \shortcite{Inp:kolar:2009:genre} and Fraser et al. \shortcite{Inp:Fraser:2015:segmentation}. We also believe that the quality of the audio recordings may have impacted the weight of the prosodic features, since the Constitution dataset was recorded by speech processing experts in a studio and the Cinderella dataset was recorded in a clinical setting. In light of this, we can see that our method performs better when all features are used. 
Furthermore, the best results were obtained by using $\alpha = 0.6$, from the linear combination in Equation~\ref{eq:linearcomb}, showing that our model lends more weight to the lexical model.

\subsection{Comparison of methods}

In order to compare our model with related work, we replicated the approach proposed by Fraser et al. \shortcite{Inp:Fraser:2015:segmentation}, which uses a CRF model for sentence segmentation. To explain the choice for a recurrent convolutional model, we split our method in three: (i) Multilayer Perceptron (MLP): we removed the convolutional and the recurrent layer of our model, and added a hidden fully-connected layer with 100 units and sigmoid activation; (ii) CNN: we simply removed the recurrent layer from our model and passed the output from the convolutional to the fully-connected layer; (iii) Recurrent Neural Network (RNN): analogous to the CNN model, we removed the convolutional layer and connected the embedding layer with the recurrent layer.
The results for each method are presented in Table~\ref{tab:results_methods}. 

\begin{table*}[!thb]
\small
\begin{center}
\begin{tabular}{lccccccccccccccc}
\toprule
\bf Methods & \multicolumn{7}{c}{\bf Cinderella} & & \multicolumn{7}{c}{\bf Constitution} \\ 					
			   \cmidrule{2-8} 
               \cmidrule{10-16}
             
               & \multicolumn{3}{c}{\bf CTL} & & \multicolumn{3}{c}{\bf MCI} & & 
                 \multicolumn{3}{c}{\bf L} & & \multicolumn{3}{c}{\bf S} \\
               
               \cmidrule{2-4} 
               \cmidrule{6-8}
               \cmidrule{10-12}
               \cmidrule{14-16}
             
             & \it P & \it R & \it $F_1$ & & \it P & \it R & \it $F_1$ & &
               \it P & \it R & \it $F_1$ & & \it P & \it R & \it $F_1$ \\  
\midrule 
CRF 	& 0.70 & 0.45 &  0.55  		& &  0.62 & 0.46 & 0.53   	 & &  0.89 & 0.36 & 0.51  	  & &  0.84 & 0.34 & 0.48  		\\
MLP 	& 0.59 & 0.79 &  0.67  		& &  0.47 & 0.80 & 0.59   	 & &  0.75 & 0.79 & 0.77  	  & &  0.76 & 0.80 & 0.78   	\\
RNN 	& 0.27 & 0.68 &  0.39  		& &  0.73 & 0.25 & 0.37   	 & &  0.43 & 0.92 & 0.58  	  & &  0.44 & 0.85 & 0.57  		\\
CNN 	& 0.64 & 0.79 &  0.71  		& &  0.59 & 0.77 & 0.67   	 & &  0.65 & 0.85 & 0.73 	  & &  0.58 & 0.89 & 0.70   	\\
RCNN	& 0.72 & 0.76 &  \bf 0.74 	& &  0.66 & 0.74 & \bf 0.70  & &  0.77 & 0.82 & \bf 0.79  & &  0.76 & 0.85 & \bf 0.80  	\\  
\bottomrule
\end{tabular}
\end{center}
\caption{\label{tab:results_methods} Best $F_1$ results for each method.}
\end{table*}

Our method achieved the best results in both datasets. We can see that the CRF method, used by Fraser et al. \shortcite{Inp:Fraser:2015:segmentation}, obtained the worst results on Constitution, and was only better than RNN on the Cinderella data. These results were similar to those reported in their paper, which suggests that our replication was faithful.
We believe that the RNN performed poorly because it has a large set of weights to be trained, and since we have relatively little data, it failed to achieve good results. This may be related to the fact that LSTM units are very complex and need more data to be able to converge. Looking at the Constitution results, which have about three times more words than the Cinderella data, we can note the difference ($\sim 0.2$) with relation to corpus size.

MLP and CNN alone were able to achieve better results than CRF and RNN, but MLP results for the MCI subset were not as good as CNN, which indicates that MLP alone is not able to deal with narratives that are potentially impaired. However, for the Constitution data, MLP obtained results very close ($\sim 0.02$) to our best method.

Our RCNN achieved the best results on both datasets, implying that a union of these models was a good choice in order to deal with impaired speech. We believe that the greatest influence was from the CNN, and the addition of a recurrent layer with LSTM was able to deal with some particular cases, likely over long dependencies similar to the findings in \cite{Inp:Tilk:2015:lstm}, where the CNN was not able to do so due to the fixed filter length in the convolution process,  a result which was also noted in \cite{Art:Che:2016:convolutional}.




\subsection{Robustness tests}

Robustness was evaluated by measuring $F_1$ on both out-of-genre and in-genre data. The results for each configuration are presented in Table \ref{tab:results_robustness}.

\begin{table}[!htb]
\small
\begin{center}
\begin{tabular}{llccc}
\toprule
\bf Trained on 		& \bf Tested on 	&  \it P  & \it R  & \bf $F_1$ 	\\
\midrule
Constitution 	& Cinderella CTL			&  0.19   	&   0.29	&   0.23   	\\
Constitution	& Cinderella MCI			&  0.20   	&   0.25	&   0.22   	\\
Cinderella		& Dog story CTL			&  0.72   	&   0.62	&   0.66   	\\
Cinderella		& Dog story MCI			&  0.65   	&   0.64	&   0.64    \\
\bottomrule
\end{tabular}
\end{center}
\caption{\label{tab:results_robustness} Results for robustness tests} 
\end{table}	

We evaluated our method by changing the corpus genre: training with the Constitution and testing with the Cinderella dataset. This evaluation shows that our method performed poorly in this scenario, probably because the differences in the lexical clues between these datasets are high, since the Constitution is composed of prepared speech and Cinderella of spontaneous speech.
When we maintain the corpus genre but change the story used in the neuropsychological test, our method can still achieve good results, yielding a small difference of $0.08$ for CTL and $0.06$ for MCI from our best results. We believe that these results are related with the linear combination weight from Equation \ref{eq:linearcomb}, where the results were obtained by using $\alpha = 0.8$, lending less weight to the prosodic model when compared to our best results (where it has $40\%$ of influence). Since the Dog Story and Cinderella datasets are composed of spontaneous speech, the lexical clues found in this kind of speech helped the method to achieve good performance.

\section{Conclusions and Future Work}


We have shown that our model, using a recurrent convolutional neural network, is benefited by word embeddings and can achieve promising results even with a small amount of data. We found that our method is better for cases where speech is planned, since the prosodic features lend more weight to the classification. 
Our method achieved good results on impaired speech transcripts even with little data, with an $F_1$ result of 0.74 on CTL patients, which is comparable with the results from other studies using broadcast news and conversational data \cite{Art:Wang:2012:dynamic,Inp:Khomitsevich:2015:combining,Inp:Tilk:2015:lstm,Art:Che:2016:convolutional}. Moreover, our method achieved good results in robustness tests when we changed the story used in the neuropsychological test.

As for future work, we plan to evaluate our method on English data for comparison with related work. Also, we plan on using more text data to train the lexical model, as it is independent from the prosodic model and lends more weight in our evaluations. Moreover, we will evaluate our method with the output of an ASR system for BP, as a higher word recognition error rate can greatly affect our results. 
Lastly, we would like to evaluate our method with datasets with higher quality audio, more robust acoustic models and a manually aligned portion of the database as better audio segmentation would greatly improve the model and the usefulness of prosodic features.

With respect to improvements in the corpus, our dataset consists of spontaneous speech narratives and was annotated only with periods. Since there are initial conjunctions such as “and”, “moreover”, and “however”, we could include commas. This would turn our problem into a ternary problem. This could be done by increasing the number of neurons in the last layer of our architecture.


\section*{Acknowledgments}

We thank CNPq for a scholarship granted to the first author.

\bibliography{eacl2017}
\bibliographystyle{eacl2017}

\end{document}